\documentclass[conference]{IEEEtran}
\IEEEoverridecommandlockouts
\usepackage{cite}
\usepackage{amsmath,amssymb,amsfonts,dsfont}
\usepackage{mathtools}

\DeclareMathOperator{\Fidelity}{Fidelity}
\DeclareMathOperator{\Consistency}{Consistency}
\DeclareMathOperator{\Prob}{Prob}
\DeclareMathOperator{\tr}{tr}
\DeclareMathOperator{\logdet}{log\:det}

\def\Lz{L$_0$}
\def\Lo{L$_1$}
\def\Loo{\Lo-O}

\input{math.tex}
\usepackage{algorithm, algorithmic}
\usepackage{graphicx}
\usepackage{subcaption}
\usepackage{wrapfig}
\usepackage{tikz}
\usepackage{multirow, multicol}
\usepackage{stfloats} 
\usepackage{url}

\usepackage{textcomp}
\usepackage{xcolor}
\def\BibTeX{{\rm B\kern-.05em{\sc i\kern-.025em b}\kern-.08em
    T\kern-.1667em\lower.7ex\hbox{E}\kern-.125emX}}
    
\begin{document}

\title{Enhancing Decision Tree based Interpretation of
	Deep Neural Networks through L1-Orthogonal
	Regularization\\
}

\author{\IEEEauthorblockN{Nina Schaaf}
\IEEEauthorblockA{\textit{Center for Cyber Cognitive Intelligence CCI} \\
\textit{Fraunhofer IPA}\\
Stuttgart, Germany \\
nina.schaaf@ipa.fraunhofer.de}
\and
\IEEEauthorblockN{Marco F. Huber}
\IEEEauthorblockA{\textit{Center for Cyber Cognitive Intelligence CCI} \\
\textit{Fraunhofer IPA}\\
\textit{Institute for }
Stuttgart, Germany \\
marco.huber@ieee.org}
\and
\IEEEauthorblockN{Johannes Maucher}
\IEEEauthorblockA{
\textit{Stuttgart Media University}\\
Stuttgart, Germany \\
maucher@hdm-stuttgart.de}
}

\author{Nina Schaaf$^{1}$, Marco F. Huber$^{1,2}$, and Johannes Maucher$^{3}$
\thanks{$^{1}$Nina Schaaf and Marco Huber are with the Center for Cyber Cognitive Intelligence (CCI), Fraunhofer Institute for Manufacturing Engineering and Automation IPA, Nobelstra{\ss}e 12, 70569 Stuttgart, Germany
{\tt\small \{nina.schaaf, marco.huber\}@ipa.fraunhofer.de}}%
\thanks{$^{2}$Marco F. Huber is with the Institute of Industrial Manufacturing and Management IFF, University of Stuttgart, Allmandring 35, 70569 Stuttgart, Germany
{\tt\small marco.huber@ieee.org}}%
\thanks{$^{3}$Johannes Maucher is with the Stuttgart Media University, 70569 Stuttgart, Germany
{\tt\small maucher@hdm-stuttgart.de}}%
}

\maketitle

\begin{abstract}
One obstacle that so far prevents the introduction of machine learning models primarily in critical areas is the lack of explainability. In this work, a practicable approach of gaining explainability of deep artificial neural networks (NN) using an interpretable surrogate model based on decision trees is presented. 
Simply fitting a decision tree to a trained NN usually leads to unsatisfactory results in terms of accuracy and fidelity. Using \Lo-orthogonal regularization during training, however, preserves the accuracy of the NN, while it can be closely approximated by small decision trees. Tests with different data sets confirm that \Lo-orthogonal regularization yields models of lower complexity and at the same time higher fidelity compared to other regularizers.
\end{abstract}


\section{Introduction}
In the recent years, machine learning (ML) gained an increasing interest in a multitude of domains like manufacturing, health care or finance. Deep learning approaches trained on large data sets are already able to compete with and even outperform humans in decision making in some applications like games\footnote{\url{https://deepmind.com/research/alphago/}, last accessed 30 Sep 2019} or medical science\footnote{\url{https://cs.stanford.edu/people/esteva/nature/}, last accessed 30 Sep 2019}. Many ML models are considered as ``black box'', i.e., decisions made are often not comprehensible to humans due to complex internal processes.
It is this complexity, however, that makes modern ML algorithms so powerful---they find patterns in large, high-dimensional data sets that no human could ever discover and the lack of comprehensibility is perfectly sufficient for some applications like movie recommendation or machine translation. For many use cases, however, there is an interest in making black-boxes transparent.

One way to enforce so-called \emph{explainability} of ML models is to create models for which the explainability objective is an essential part of their design. \emph{Ante-hoc} models are designed to be inherently explainable; examples for this type of models are logistic regression, rule-based systems~\cite{Wang.2017}, or decision trees. In contrast, \emph{post-hoc} explainability refers to adding the explainability objective after training~\cite{Ribeiro.20160809,Zilke.2016}. Furthermore, model explainability techniques can either be applicable to a specific model type (\emph{model-specific}) or to a variety of different models (\emph{model-agnostic}).
When searching for a suited approach, one has to additionally define the scope of the generated explanations. \emph{Global} explainability---sometimes also called model explainability---implies understanding the model as a whole. Global explainability enables, for example, insights into non-linearities or feature interactions.
In contrast, \emph{local} or output explainability entails knowing the reasons for a specific prediction (or a group of predictions) rather than examining the entire model~\cite{Molnar.2018}.

Depending on the desired goal, information on the inner workings of ML models can be represented in various ways, for example using visual representations such as heat-maps~\cite{Bach.2015} or plots~\cite{Friedman.2001}. Another possibility towards providing explainability is rule extraction from black-box models~\cite{Lakkaraju.2017,Puri.2017,Zilke.2016}. 
Rule extraction techniques for NNs follow the idea of deriving simple human comprehensible rules from NNs in order to approximate the network's decision-making process and hence provide explanation capability. Information extracted from NNs can be represented in different formats such as decision trees, decision tables or simple rules of the form:
\begin{quote}
\texttt{If weather=sunny and windspeed=low then play tennis}
\end{quote}
In this paper simple but deep NNs named multi-layer perceptrons (MLP) are considered. 
A number of works have been published that focus on rule extraction from MLPs. For example, DeepRED~\cite{Zilke.2016} uses decision trees as an instrument to extract rules from deep MLPs at the level of the networks' individual neurons. RxREN~\cite{Augasta.2011} prunes insignificant input neurons from a trained MLP and identifies data ranges for the remaining input neurons for rule extraction.

Instead of touching the inner structure of a trained MLP in a post-hoc fashion, the focus in this paper is on \emph{optimizing} deep MLPs towards post-hoc decision tree extraction. For this purpose, a combination of L$_1$ and orthogonal regularization is proposed, which favors MLPs with decision boundaries that can be approximated much easier by small decision trees. These   models allow explainability on a \emph{global} level. By means of numerical evaluation on various public benchmark datasets it is shown that the L1-orthogonal regularization approach yields MLPs that maintain a high accuracy, while the extracted decision tress achieve a high fidelity, i.e., they approximate the MLPs well and thus, provide consistent, meaningful, and human comprehensible explanations for deep models.

\section{Problem Statement}
\label{sec:problem}
Multi-class classification is considered, where training is performed by means of a labeled data set $\SD = \{\vx_n, y_n\}_{n=1}^N$ comprising $N$ samples, where $\vx_n \in \NewR^d$ is the input vector of the $n$-th sample and $y_n \in \NewN$ is the corresponding label. The elements of $\vx_n$ are named features or attributes.

\subsection{Multi-layer Perceptron (MLP)}
To learn the (unknown) mapping from input $\vx$ to output $y$, an MLP is employed, which consists of multiple neurons that are arranged in two or more layers. These neurons are linked via weighted connections, where the weights of connections from the neurons in layer $l-1$ to those in layer $l$ are stored in a weight matrix $\mat W_l$, with $l \in \{1,...,L\}$. In the following, $\SW = \{\mat W_l\}_{l=1}^L$ is the collection of all weight matrices. A column $\vw_i$ of a weight matrix $\mat W_l$ comprises the weights of connections 
from all neurons of layer $l-1$ to the $i$-th neuron of layer $l$.

An MLP's output $\hat{y}_n$ is computed via a function $f(\vx_n, \SW)$. MLP training aims for adjusting the weights in $\SW$ by means of solving the optimization problem 
\begin{equation}
\label{equ:loss}
\min_\SW\ \sum_{n=1}^{N} E\big(y_n, f(\vx_n, \SW)\big) + \lambda \cdot \Omega(\SW)
\end{equation}
such that the error $E$ between the MLP's outputs $\hat{y}_n$ and the given targets $y_n$ is minimized. In \eqref{equ:loss}, $\Omega(\SW)$ is a so-called \emph{regularization term}, whose strength can be controlled by a regularization parameter $\lambda \in \NewR^+$. Usually, this term is used to avoid overfitting.

\subsection{Decision Trees (DT)}
In this paper, DTs are used for extracting explanations from a trained MLP. 
A DT consists of internal and leaf nodes. The internal nodes specify tests of the value of one of the input attributes whereas the leaf nodes specify class labels~\cite{Russell.2010}. To classify an example, the tree is traversed from top to bottom. The branches of an internal node correspond to the outcomes of the test. Depending on the test outcome, the corresponding branch is followed. This procedure is repeated at each of the following internal nodes until a leaf node is reached. Every path from the tree's root to a leaf node can be translated into a if-then-rule like the one shown above. The condition of the if-clause then corresponds to the tests along the path.

\subsection{Related Work on DT Extraction}
In Alg.~\ref{alg:l1-ortho-dt-extraction}, a meta-algorithm for DT extraction from a trained MLP is shown. Here, the MLP is considered as an oracle that is fed with various inputs $\vx$ to predict outputs $y$. This data is then used to learn a DT. A na\"ive way is to use no regularization, i.e., $\Omega \equiv 0$. In \cite{Craven.1994.b} this approach was extended by considering the data distribution and constraints when sampling the inputs, but MLP training is not influenced to improve the accuracy of the extracted DT.

\begin{algorithm}[tbp]
	\caption{MLP training and subsequent DT extraction. Given a labeled data set $\SD$ and regularizer $\Omega$, the MLP is trained by solving \eqref{equ:loss} together with an appropriate loss function $E$. The resulting weight matrices $\SW$ and the training data $\{\vx_n\}_{n=1}^N$ are used to generate predictions $\hy_n$ from the MLP. Finally, the DT is trained with $\{\vx_n, \hy_n\}_{n=1}^N$.}
	\label{alg:l1-ortho-dt-extraction}
	\begin{algorithmic}
		\renewcommand{\algorithmicrequire}{\textbf{Input:}}
		\renewcommand{\algorithmicloop}{\textbf{function} \textsc{TrainAndExtractDT}:}
		\renewcommand{\algorithmicendloop}{\algorithmicreturn}
		\REQUIRE $\SD=\{\vx_n, y_n\}_{n=1}^N$: data set with $N$ examples,\\
		\hspace{8mm}$\Omega(.)$: regularization term
		\LOOP 
		\STATE $\SW \longleftarrow$ \textsc{TrainMLP}$(\SD, \Omega)$
		\STATE $\{\hy_n\} \longleftarrow f(\{\vx_n\},\SW)$
		\STATE tree $\longleftarrow$ \textsc{TrainDT}$\big(\{\vx_n, \hy_n\}\big)$
		\ENDLOOP
		\ $\SW$, tree
	\end{algorithmic}
\end{algorithm}

Wu et al.~\cite{MikeWu.2018} instead present a novel regularizer called \emph{tree regularization}, where $\Omega$ represents the \emph{average-path-length} (APL) cost function. The idea is to train an MLP in such a way that the extracted DT are rather shallow. Unfortunately, APL cannot be calculated in closed-form: a DT needs to be trained first in order to calculate the APL. Thus, APL is also not differentiable, which complicates MLP training in addition. To overcome these issues, \cite{MikeWu.2018} propose to learn an additional surrogate MLP for estimating the APL. It is obvious that simultaneously training two interrelated MLPs requires significant training time and careful parameter tuning.

\section{L1-Orthogonal Regularization}
\label{section:l1-ortho-reg}
The goal is to avoid the limitations of tree regularization but at the same time maintain its appealing idea. For this purpose a closed-form, simple to implement, and differentiable regularization is proposed that forces the decision boundaries of the trained MLP to be easily approximated by a DT. In addition, the trained MLP should result in a DT with shorter path lengths than a DT extracted from an MLP trained without or with other regularizers. However, the MLP's predictive accuracy should not decrease significantly when applying the regularizer during training. In order to achieve this goal, we propose promoting both sparseness and orthogonality of the weight vectors $\vw$ in the weight matrices $\W$. 

It is well known that a weight vector represents the normal vector of a linear decision boundary. By means of enforcing a sparse representation, where many or even all but one elements of a weight vector are close to zero, the linear decision boundary becomes axis parallel. This representation is likely to harmonize well with DTs. It can be shown that the decision boundary of a DT consists of axis-parallel segments, as each test node in a DT divides the input space into axis-aligned hyperplanes, where each hyperplane is labeled with one class. 

Further, we want to avoid that too many decision boundaries are (almost) parallel to each other, i.e., many weight vectors point into a similar direction. This would limit the predictive power of the model and thus, sparseness is combined with orthogonality, i.e., the weight vectors are encouraged to be close to orthogonal in a pair-wise fashion. By combining sparse with orthogonal regularization during MLP training it is intended that a weight matrix $\W$ contains a small number of non-zero entries (sparse) that nevertheless cover a broad spectrum of features (orthogonal). This kind of regularization drives MLPs to have decision boundaries being more similar to those of DTs and thus, can be better approximated. 

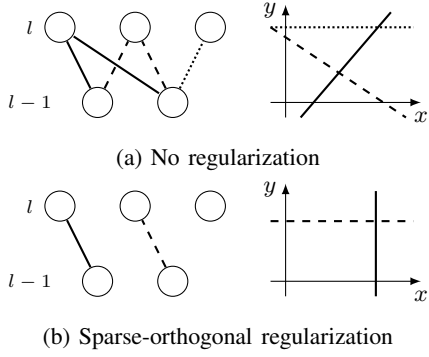
\begin{figure}[tb]
	\centering
	\begin{subfigure}[b]{0.49\textwidth}
		\centering
		\begin{tikzpicture}
        	\draw (.5,0) circle (2mm) node(ur){};
        	\draw (-.5,0) circle (2mm) node(ul){};
        		
        	\draw (-1,1) circle (2mm) node(ol){};
        	\draw (0,1) circle (2mm) node(om){};
					\draw (1,1) circle (2mm) node(or){};
        	
        	\node at (-1.4,0) {\scriptsize $l-1$};
        	\node at (-1.4,1) {\scriptsize $l$};
        	
        	\draw[thick, shorten <= 2, shorten >= 2] (ul) -- (ol);
        	\draw[thick, shorten <= 2, shorten >= 2] (ur) -- (ol);
        	\draw[thick, dashed, shorten <= 2, shorten >= 2] (ul) -- (om);
        	\draw[thick, dashed, shorten <= 2, shorten >= 2] (ur) -- (om);
        	\draw[thick, densely dotted, shorten <= 2, shorten >= 2] (ur) -- (or);
        	
        	\draw[-latex] (1.8,0) -- (3.8,0) node[at end, below] {\small $x$};
        	\draw[-latex] (2.0,-.2) -- (2.0,1.3) node[pos=.95, left] {\small $y$};
        	\draw[thick] (2.2,-.2) -- (3.4,1.2);
        	\draw[thick, dashed] (1.8,1.0) -- (3.6,-.2);
        	\draw[thick, densely dotted] (1.8,1.0) -- (3.7,1.0);
        \end{tikzpicture}
		\caption{No regularization}
		\label{fig:no-regularization}
	\end{subfigure}
	\begin{subfigure}[b]{0.49\textwidth}
		\centering
		\begin{tikzpicture}
        	\draw (.5,0) circle (2mm) node(ur){};
        	\draw (-.5,0) circle (2mm) node(ul){};
        		
        	\draw (-1,1) circle (2mm) node(ol){};
        	\draw (0,1) circle (2mm) node(om){};
					\draw (1,1) circle (2mm) node(or){};
        	
        	\node at (-1.4,0) {\scriptsize $l-1$};
        	\node at (-1.4,1) {\scriptsize $l$};
        	
        	\draw[thick, shorten <= 2, shorten >= 2] (ul) -- (ol);
        	\draw[thick, dashed, shorten <= 2, shorten >= 2] (ur) -- (om);
        	
        	\draw[-latex] (1.8,0) -- (3.8,0) node[at end, below] {\small $x$};
        	\draw[-latex] (2.0,-.2) -- (2.0,1.3) node[pos=.95, left] {\small $y$};
        	\draw[thick] (3.2,-.2) -- (3.2,1.2);
        	\draw[thick, dashed] (1.8,.8) -- (3.7,.8);
        \end{tikzpicture}
		\caption{Sparse-orthogonal regularization}
		\label{figure:l1-orth-idea-orth}
	\end{subfigure}
	\caption{Effect of sparse-orthogonal regularization on an MLP's weight vectors.}
	\label{figure:l1-orth-idea}
\end{figure}

Figure~\ref{figure:l1-orth-idea} shows the intended effects on weight vector alignment when applying the proposed sparse-orthogonal regularization approach. Here, the weights between two consecutive layers are schematically illustrated. The weights on the connections of all neurons of layer $l-1$ to a neuron of layer $l$ form the elements of the weight vector. No connection means that the corresponding weight is equal to zero. If a neuron in layer $l$ has more than one connection, the weight vector is not axis parallel as can be seen in Fig.~\ref{fig:no-regularization}. The network in Fig.~\ref{figure:l1-orth-idea-orth} instead is both sparse (few connections) and has orthogonal weight vectors (connections to different neurons in layer $l-1$).

\subsection{Regularizers}
\label{subsection:regularizers}
In the following, different regularization terms are introduced for inducing sparsity and orthogonality into an MLP for tree extraction.

\subsubsection{L$_1$ Regularization}
Ideally, one should apply the \Lz-norm in order to obtain maximum sparsity, i.e., all elements of a column vector $\vw$ of $\SW$ are exactly zero except for one. This approach, however, is not feasible in practice, as the \Lz-norm is known to be non-differentiable. Instead, an approximation is required. A well known approximation is the \Lo-norm
\begin{equation}
\label{eq:l1}
    \Omega_1(\SW) = \textstyle\sum_l\ \lVert \W_l\rVert_1~,
\end{equation}
where $\Vert\A\rVert_1 = \sum_i \lvert \va_i\rvert = \sum_{i}\sum_j |a_{ij}|$ is the so-called \emph{\Lo-norm} of matrix $\A$, with $\va_i$ being the $i$-th column vector of~$\A$\,.

\subsubsection{Orthogonal Regularization}\label{subsubsection:ortho-reg}
The aim of orthogonal regularization is to orthogonally align the weight vectors $\vw_i$ in each layer of an MLP. 
Xie et al.~\cite{Xie.2017} propose the following method to promote orthogonality: Two vectors $\vw_i$ and $\vw_j$ are orthogonal if their inner product $\vw_i\T\vw_j$ is zero and their L$_2$-norms $\lVert\vw_i\rVert_2$ and $\lVert\vw_j\rVert_2$ are close to one. 
For simultaneously promoting orthogonality between all pairs of weight vectors of $\W$, one can consider the corresponding Gram matrix $\G_{ij} = \vw_i\T\vw_j$. The Gram matrix consists of the pair-wise scalar products of the weight matrix' columns, where the Gram matrix' diagonal elements (the Gram determinant) are $\lVert\vw_i\rVert_2$
. Encouraging $\G$ to be close to the identity matrix $\I$ makes $\vw_i\T\vw_j$ close to zero and $\lVert\vw_i\rVert_2$ close to one, resulting in near-orthogonality of the weight vectors~\cite{Xie.2017b}.

To compute the scalar product of all weight vectors of a layer's weight matrix $\W_l$ one can simply compute the Gram matrix $\G_l = \W_l\T\W_l$. 
Thus, near-orthogonality among the weight vectors in each layer $l$ can be promoted by substituting $\Omega$ in \eqref{equ:loss} with the regularization term 
\begin{equation}
\label{eq:ortho-reg}
\Omega_\mathrm{orth}(\SW) = \textstyle\sum_l\ \lVert\G_l - \I\rVert_1~.
\end{equation}
Here, $\lVert \cdot\rVert_1$ is the \Lo-norm as in \eqref{eq:l1}.

\subsubsection{Alternatives} 
As with vectors, there are several options to compute the norm of matrices. One alternative approach of measuring the closeness between $\G$ and $\I$ is using the so-called \emph{Frobenius norm} (FN)
\begin{equation}
\label{eq:frobenius}
    \lVert \A\rVert_F^2 = \textstyle\sqrt{\sum_{i}\sum_j |a_{ij}|^2}
\end{equation}
as described in~\cite{Xie.2017}. Experimental evaluations indicate (see Sec.~\ref{section:evaluation} for details) that both norms often provide similar results, with slight advantages for the \Lo-norm. Also from a numerical perspective, the \Lo-norm turned out to be more reliable and thus, we recommend to use the \Lo-norm.

A further alternative is the so-called log-determinant divergence (LDD) for measuring the closeness between $\G$ and $\I$ by encouraging their LDD 
\begin{equation}
\label{eq:ldd}
\Omega_\mathrm{LDD}(\SW) = \tr(\mathbf{G}) - \logdet(\mathbf{G}) 
\end{equation}
to be small~\cite{Xie.2017b}, where tr($\cdot$) is the matrix trace and det($\cdot$) is the matrix determinant.

However, applying \eqref{eq:ldd} to gradient descent optimization leads to the following problem. Let $f(\A)=\logdet \A$\,, the gradient is defined as $\nabla f(\A)=\A^{-\text{T}}$~\cite{Boyd.2004}. Inverting $\A$, however, is difficult to impossible when $\A$ is close to be singular. Consequently, numerical errors occur during training when the Gram matrix determinant approaches zero.

\subsubsection{\Lo-Orthogonal (\Lo-O) Regularization}
Based on the aforementioned discussions, L$_1$ regularization \eqref{eq:l1} and orthogonal regularization~\eqref{eq:ortho-reg} are combined to form \emph{\Lo-orthogonal} (\Lo-O) regularization 
\begin{equation}
    \Omega(\SW) = \lambda_1 \cdot \Omega_1(\SW) + \lambda_\mathrm{orth} \cdot \Omega_\mathrm{orth}(\SW)
\end{equation}
to be employed in \eqref{equ:loss}. But instead of a single regularization parameter $\lambda$ as in \eqref{equ:loss}, the regularization strengths of $\Omega_1$ and $\Omega_\mathrm{orth}$ are regulated independently via $\lambda_1$ and $\lambda_\mathrm{orth}$, respectively. With the independent parameterization of the regularization strengths, both regularizers and their trade-offs can be individually controlled, which turned out to be highly beneficial during empirical investigations. 

\subsection{Remark}
It is worth mentioning that L$_1$ is not new to NN training and also the combination of sparseness and orthogonality has already been proposed (see e.g., \cite{Xie.2017}). In contrast to \cite{Xie.2017}, the regularization combination proposed in this paper is numerically by far more stable as inverting singular weight matrices is avoided. But more important, to the best of our knowledge, this is the first time that sparse-orthogonal regularization has been proposed and investigated for the purpose of extracting explainable ML models from deep NNs.

\section{Experimental Setup}
To evaluate the \Lo-O regularization approach, detailed experiments are performed on simulated and real-world data sets. For this purpose, \Lo-O regularization's performance is compared with three alternative regularizers: \Lo, orthogonal and tree regularization. Additionally, as a baseline, a standalone DT classifier, i.e., a DT directly trained on the data, and a MLP without regularization are considered.

In order to additionally compare the effect of the matrix norm used to enforce orthogonality, the regularizers \eqref{eq:ortho-reg} and \eqref{eq:frobenius} are evaluated as well and are abbreviated \Lo-norm and FN, respectively, in the following.

\subsection{Evaluation Criteria}
In 1995, Andrews et al.~\cite{Andrews.1995} introduced a taxonomy consisting of five criteria for categorizing rule extraction techniques. For the evaluation, the fourth and the fifth criterion, namely the \emph{quality} of the rules and the \emph{complexity} of the rule extraction technique itself are of interest. Andrews et al. define the four quality measures for rules: 1) accuracy, 2) fidelity, 3) consistency, and 4) comprehensibility. The evaluation measures listed below are mainly based on these criteria.

\subsubsection{Complexity vs. predictive performance} Complexity of a DT is measured by the APL metric, i.e, the average number of decision nodes that must be passed to make a prediction, proposed in \cite{MikeWu.2018}. If a DT would be converted to if-then rules, the DT's would correspond to the average rule antecedent length. Thus, the APL metrics is a good indicator of the DT's complexity and consequently of its comprehensibility. The predictive performance of the DT's associated MLP is measured by the area under the ROC curve~(AUC).
\subsubsection{Fidelity} This criterion quantifies how well the DT mimics the behavior of the MLP from which it was extracted~\cite{Andrews.1995}. Fidelity is defined as the percentage of test examples on which the prediction made by a DT agrees with the prediction of the MLP~\cite{Craven.1994.b}, i.e.,
\begin{equation}
    \Fidelity = 1-\Prob\{f_\mathrm{DT} \neq f_\mathrm{N}\,|\,\ST\}~,
\end{equation}
where $\ST$ is the test set and $f_\mathrm{DT}$, $f_\mathrm{N}$ are the functions implemented by the DT and MLP, respectively. The higher the fidelity, the better the DT reflects the MLP and the more reliable are the conclusions drawn from it.
\subsubsection{Comprehensibility} Complexity vs. predictive performance measures how reduction in model complexity (measured by the APL) effects the predictive performance. The intention behind the comprehensibility criterion is to evaluate the DTs' comprehensibility by visually examining the trees themselves.
\subsubsection{Consistency} Andrews et al.~\cite{Andrews.1995} define that consistency is given if the rules extracted under different training sessions produce the same classifications of test examples. Consistency is defined as
\begin{equation}
\Consistency = \Prob\{f_{\mathrm{DT}_1} = \ldots = f_{\mathrm{DT}_S}\,|\,T\}~,
\label{eq:consistency}
\end{equation}
where $f_{\mathrm{DT}_i}$ is the function implemented by the DT extracted from the MLP in training session $i \in \{1,..., S\}$.
\subsubsection{Computational complexity} The computation times of \Lo-O and tree regularization are compared to assess the computational efficiency of the less complex \Lo-O regularization.

\subsection{Data Sets and Networks}
\subsubsection{Toy Data Set}
\begin{figure}[tb]
	\centering
	\includegraphics[width=0.225\textwidth]{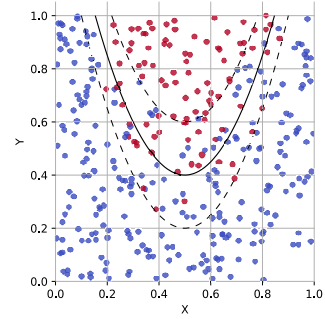}
	\caption[Training data and class labels of 2D-parabola problem.]{Training data and class labels for 2D-parabola problem.}
	\label{figure:2D-parabola}
	\vspace{-2mm}
\end{figure}
For a first comparison of the different regularizers, the 2D-parabola problem introduced in~\cite{MikeWu.2018} is chosen. As shown in Fig~\ref{figure:2D-parabola}, the training data consists of 2D input points whose two-class decision boundary is roughly shaped like a parabola and is defined by $y=5 \cdot (x-0.5)^2+0.4$\,. For data generation, 500 points are sampled uniformly within a box of size $[0,1]\times[0,1]$. Those points lying above the decision boundary are labeled positive. Finally, 10\% of the points near the boundary (limited by $y=5 \cdot (x-0.5)^2+0.2$ and $y=5\cdot (x-0.5)^2+0.6$, plotted as dashed lines in Fig.~\ref{figure:2D-parabola}) are flipped. The data set is split into subsets for training (70\%) and testing (30\%).

The MLP trained on the parabola data comprises three hidden layers with 100, 100, and 10 neurons, respectively. As this architecture encourages overfitting on this relatively small data set, this should emphasize the effects of regularization. The tree regularization's surrogate MLP has one hidden layer with 25 nodes. The main MLP applies ReLU activation in the hidden layers and sigmoid activation in the output layer while the surrogate MLP applies Tanh activation in the hidden layer and softplus activation in the output layer. The objective in~\eqref{equ:loss} is optimized via Adam gradient descent~\cite{Kingma.2014}. 

\subsubsection{Real-World Data Sets}
The 2D-parabola problem was chosen to compare \Lo-O regularization's performance to tree regularization on a simple data set. Additionally, further experiments with more complex data sets were performed. These data sets cover different sizes (150--100,000) and varying numbers of features (4--50). With these data sets, both shallow and deep MLPs were trained---deep MLPs are defined as networks comprising more than one hidden layer.

\begin{table*}[t!]
	\center
	\caption{Data sets and training parameters used for evaluation. \#neurons refers to number of neurons per hidden layer.
	}
	\label{table:evaluation:data-sets}
	\begin{tabular}{l|ccc|cccc|cc}
		\textbf{Data set} & \textbf{instances}  & \textbf{features}  & \textbf{classes} & \textbf{\#neurons} & \textbf{batch size} & \textbf{lr} & \textbf{epochs} & \textbf{min samples leaf} & \textbf{pruning}
		\\ \hline
		2-D parabola & 500 & 2 & 2 & 100,\,100,\,10 & 100 & 0.001 & 1000 & - & N\\
		Iris~\cite{Fisher.1936} & 150 & 4 & 3 & 8 & 10 & 0.01 & 50 & 5 & Y\\
		Breast Cancer Wisconsin~\cite{Dheeru.2017} & 569 & 30 & 2 & 64, 32 & 10 & 0.001 & 10 & 15 & Y\\
		Pima Indians Diabetes~\cite{Dheeru.2017} & 768 & 8 & 2 & 24 & 128 & 0.01 & 10 & 30 & Y\\
		Titanic~\cite{KaggleInc..2012} & 891 & 11 & 2 & 100,\,50,\,25 & 16 & 0.005 & 10 & 35 & Y\\
		Mushroom~\cite{Dheeru.2017} & 8,124 & 22 & 2 & 16 & 10 & 0.005 & 25 & 45 & Y\\
		Adult~\cite{Dheeru.2017}  & 48,842 & 14 & 2 & 32, 16 & 32 & 0.005 & 10 & 75 & Y\\
		Diabetes~\cite{Strack.2014}   & 100,000 & 50 & 2 & 32, 16, 8 & 512 & 0.01 & 50 & 250 & Y\\
	\end{tabular}
	\vspace{-2mm}
\end{table*}


Table~\ref{table:evaluation:data-sets} summarizes the data sets and sizes of the hidden layers of the MLPs used for the experiments. \emph{Iris} is a multi-class data set and distinguishes three iris plants. \emph{Breast Cancer Wisconsin (Diagnostic)} data set contains information on cell nuclei that are used to determine whether cancer is malignant or benign, while with the \emph{Pima Indians Diabetes} data set one can investigate whether a patient shows signs of diabetes. The \emph{Titanic} data set contains information on passengers of the Titanic and whether they survived the sinking of the ship. The \emph{Mushroom} data set describes mushrooms in terms of physical characteristics and distinguishes between poisonous and edible mushrooms. Using the \emph{Adult} data set, that contains census data, the prediction task is to determine whether a person's income exceeds US\$\,50,000 per year. \emph{Diabetes} data set represents clinical care at 130 US hospitals and integrated delivery networks during 1999--2008
and is used to predict whether a patient gets readmitted to hospital.

The MLPs listed in Table~\ref{table:evaluation:data-sets} apply ReLU activation in the hidden layers and sigmoid and softmax activation in the output layer for binary and multi-class classification, respectively. The MLPs' objective was optimized via Adam gradient descent~\cite{Kingma.2014}. Hyperparameters (layer sizes, batch size, epochs, learning rate) were set via 5-fold cross validation~(CV) using grid search. For tree regularization the same architecture for the surrogate MLP as with the toy data is used.

\subsection{Training}
For MLP training the input features $\vx$ are standardized to have zero mean and unit variance. 
Each data set is split into subsets for training (60\%), validation (20\%) and testing (20\%). Afterwards, an extensive search is carried out using a wide range of regularization parameters for each regularization type ($\lambda_1 = [0.001, 0.1], \lambda_{orth}=[0.0001,2.0]$) . In doing so, it is possible to cover as many decision boundary complexities as possible for each technique. All weights are initialized using the same random seed.

DT training is carried out using scikit-learn's~\cite{Buitinck.2013} \texttt{DecisionTreeClassifier} module. Each DT is trained with $\SD'=\{\vx, f(\vx', \mathcal{W})\}$~, where $\vx'$ are the standardized input features and $f(\vx', \mathcal{W})$ is the MLP's prediction on these features. In order to generate a smoother tree, DTs are trained with a fixed \emph{min\_samples\_leaf} parameter that defines the minimum number of samples required to form a leaf node~\cite{Buitinck.2013}. Just like the hyperparameters of the MLP, the fixed value \emph{min\_samples\_leaf} is determined via 5-fold CV. After training, the post-pruning algorithm described in~\cite{MikeWu.2018} is applied to additionally simplify the tree, using the held-out validation~set.

The same procedure applies for DTs without an associated MLP---trained on the original data set $\SD$. Additionally, in order to produce trees of different sizes, the training is performed with a) no \emph{max\_depth} parameter and b) \emph{max\_depth} parameters from 1--10.

Since training tree-regularized MLP's turned out to be problematic in terms of achieving APL reduction for increasing regularization strengths, it is only possible to provide results for 2D-parabola, Iris, and Breast Cancer data sets. One explanation for the challenging training process could be the high amount of free parameters (for the main MLP as well as the surrogate MLP), which leads to a wide spectrum of potential error sources.

\section{Experimental Results}\label{section:evaluation}
\subsection{Complexity vs. predictive performance}
\begin{figure*}[th]
	\begin{subfigure}[t]{0.49\textwidth}
		\centering
		\includegraphics[width=\textwidth]{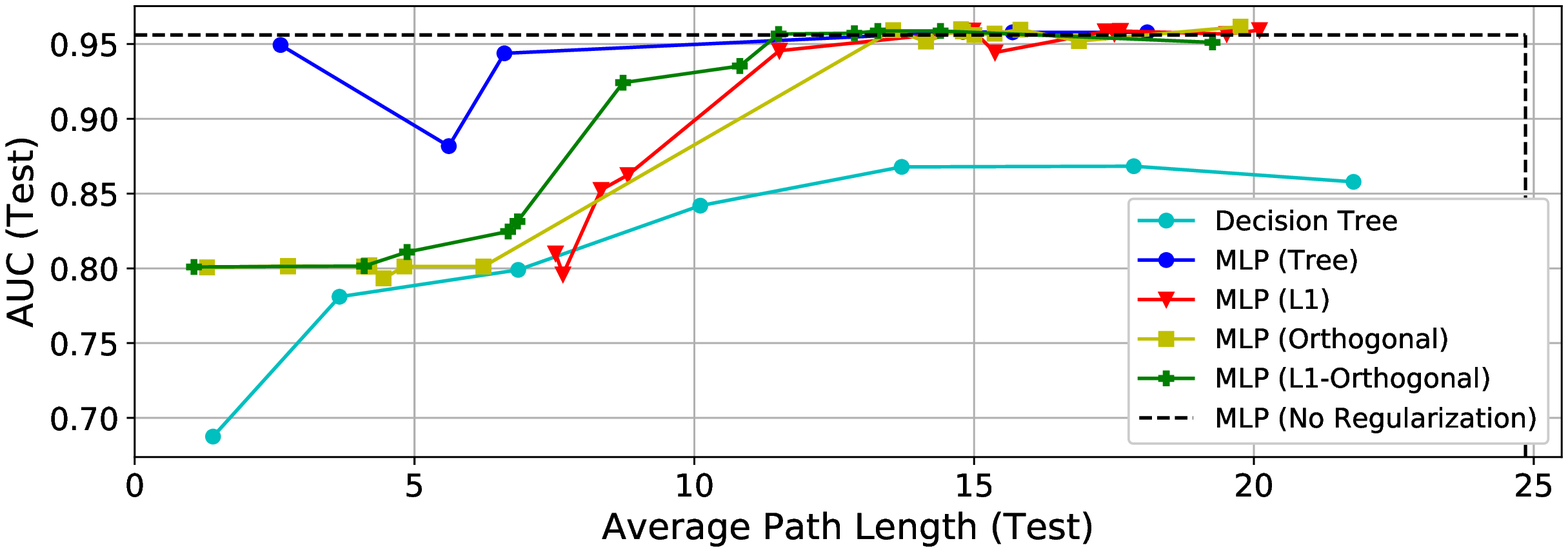}
		\vspace{-6mm}
		\caption{2D-Parabola}
	\end{subfigure}
	\begin{subfigure}[t]{0.49\textwidth}
		\centering
		\includegraphics[width=\textwidth]{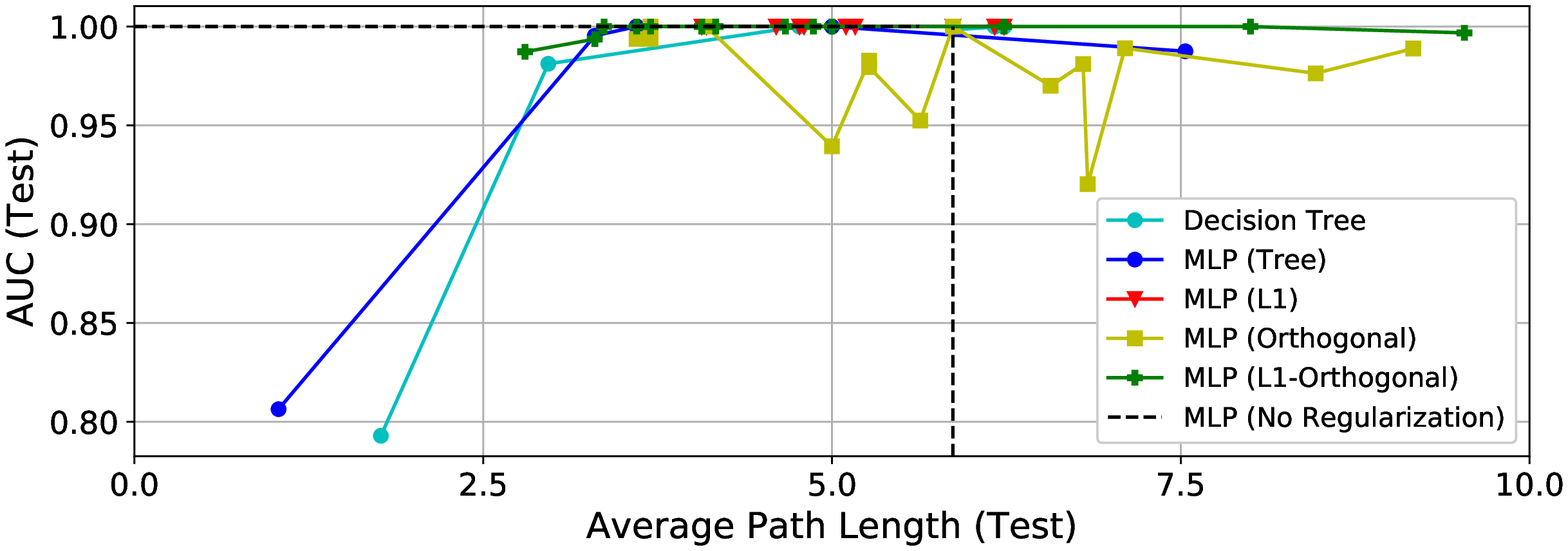}
		\vspace{-6mm}
		\caption{Iris}
	\end{subfigure}
	\vspace{-.5mm}
	\begin{subfigure}[t]{0.49\textwidth}
		\centering
		\includegraphics[width=\textwidth]{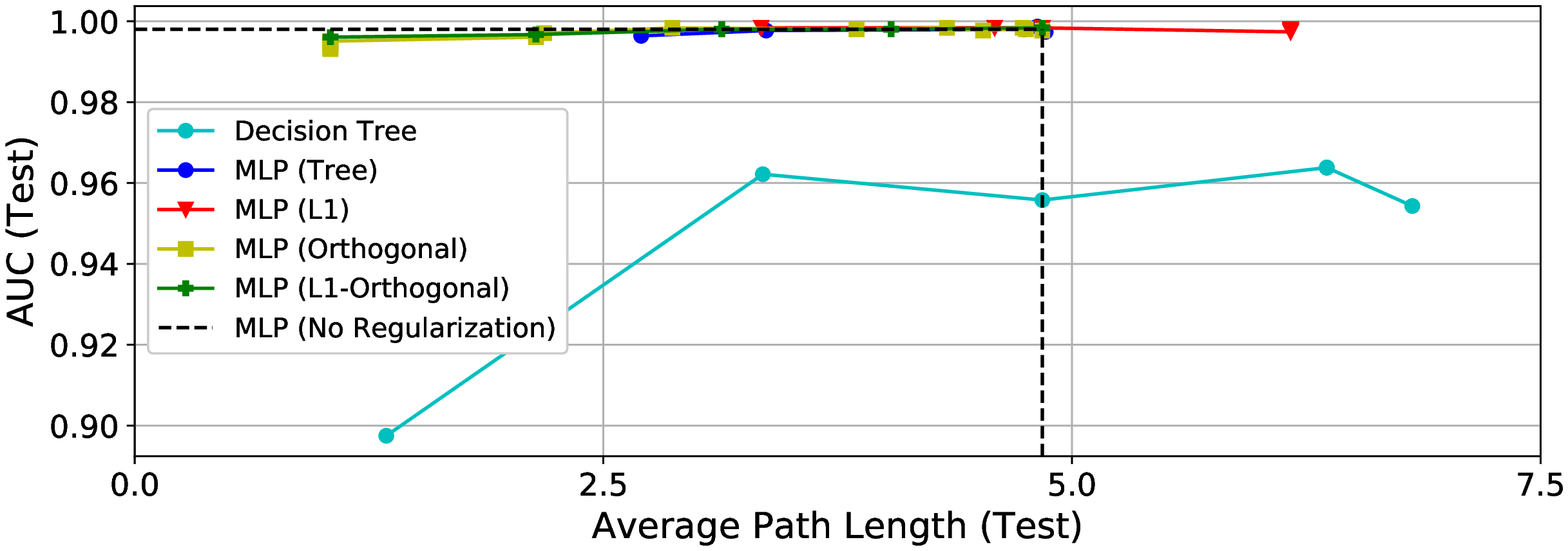}
		\vspace{-6mm}
		\caption{Breast Cancer Wisconsin}
	\end{subfigure}
	\begin{subfigure}[t]{0.49\textwidth}
		\centering
		\includegraphics[width=\textwidth]{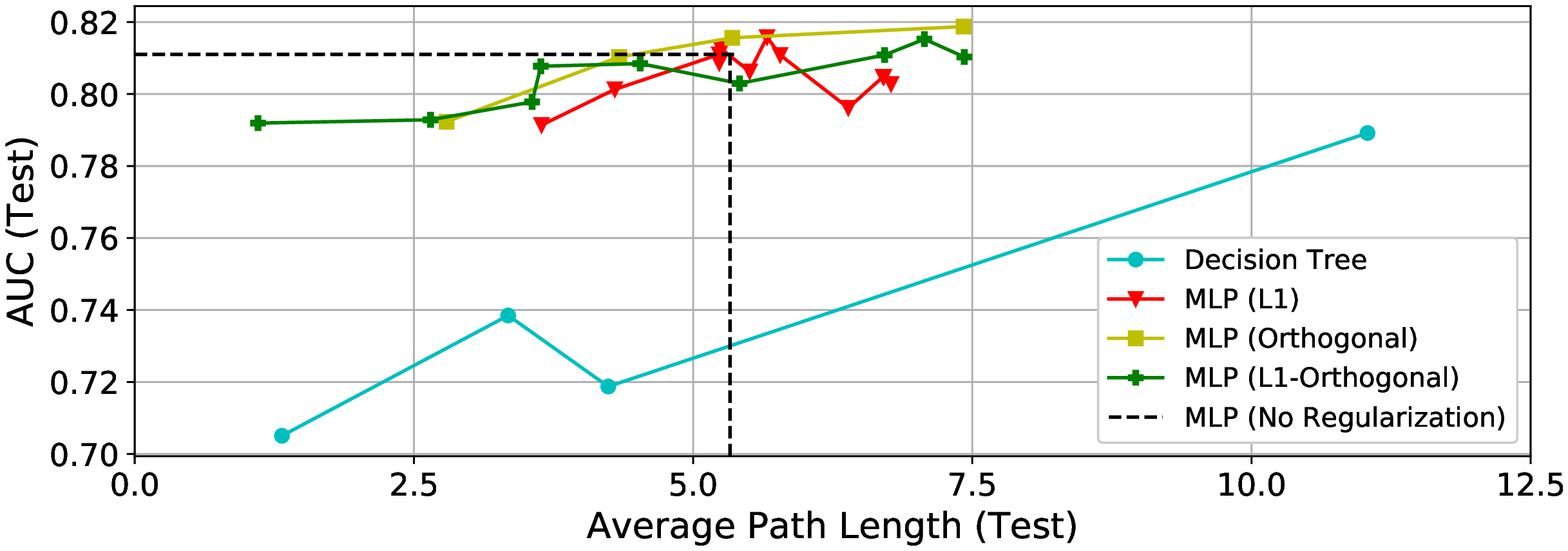}
		\vspace{-6mm}
		\caption{Pima Indians Diabetes}
	\end{subfigure}
	\vspace{-.5mm}
	\begin{subfigure}[t]{0.49\textwidth}
		\centering
		\includegraphics[width=\textwidth]{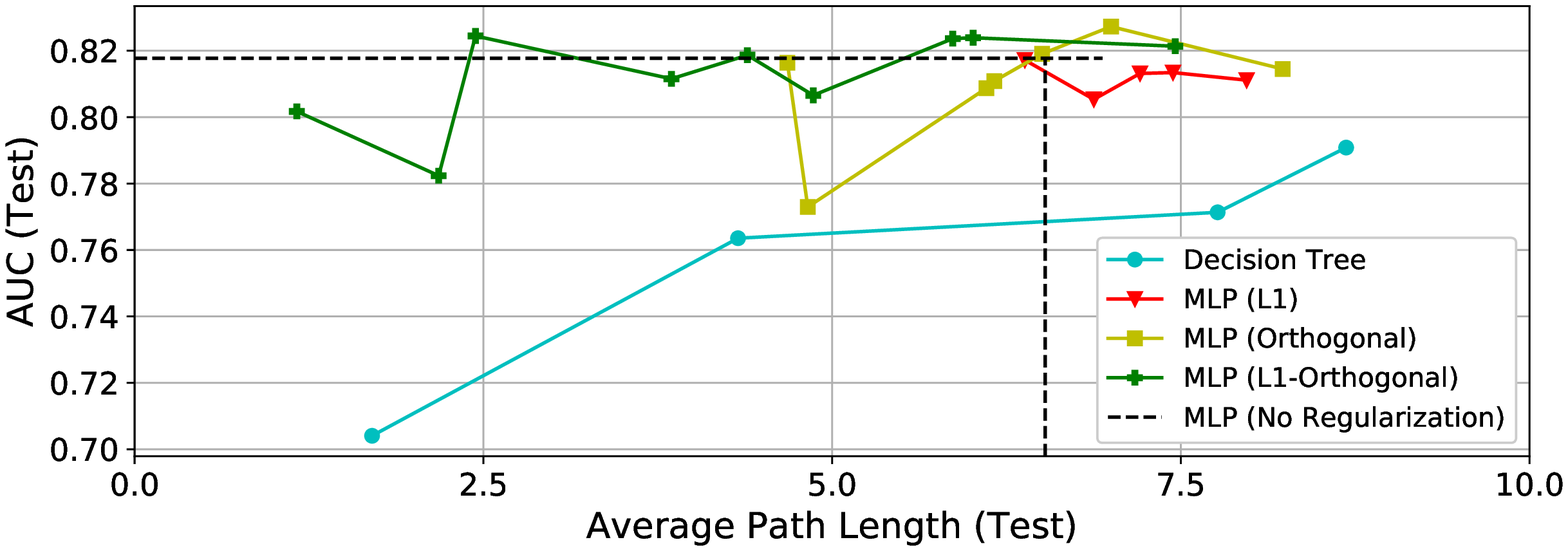}
		\vspace{-6mm}
		\caption{Titanic}
	\end{subfigure}
	\begin{subfigure}[t]{0.49\textwidth}
		\centering
		\includegraphics[width=\textwidth]{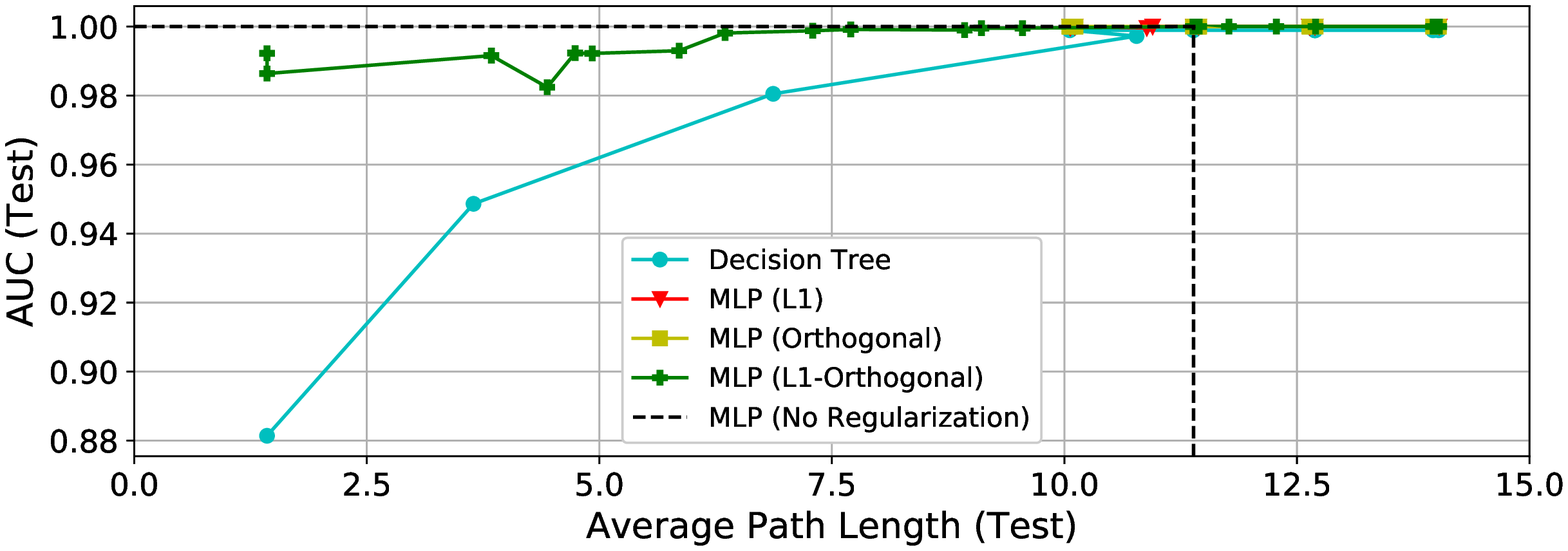}
		\vspace{-6mm}
		\caption{Mushroom}
	\end{subfigure}
	\vspace{-1mm}
	\begin{subfigure}[t]{0.49\textwidth}
		\centering
		\includegraphics[width=\textwidth]{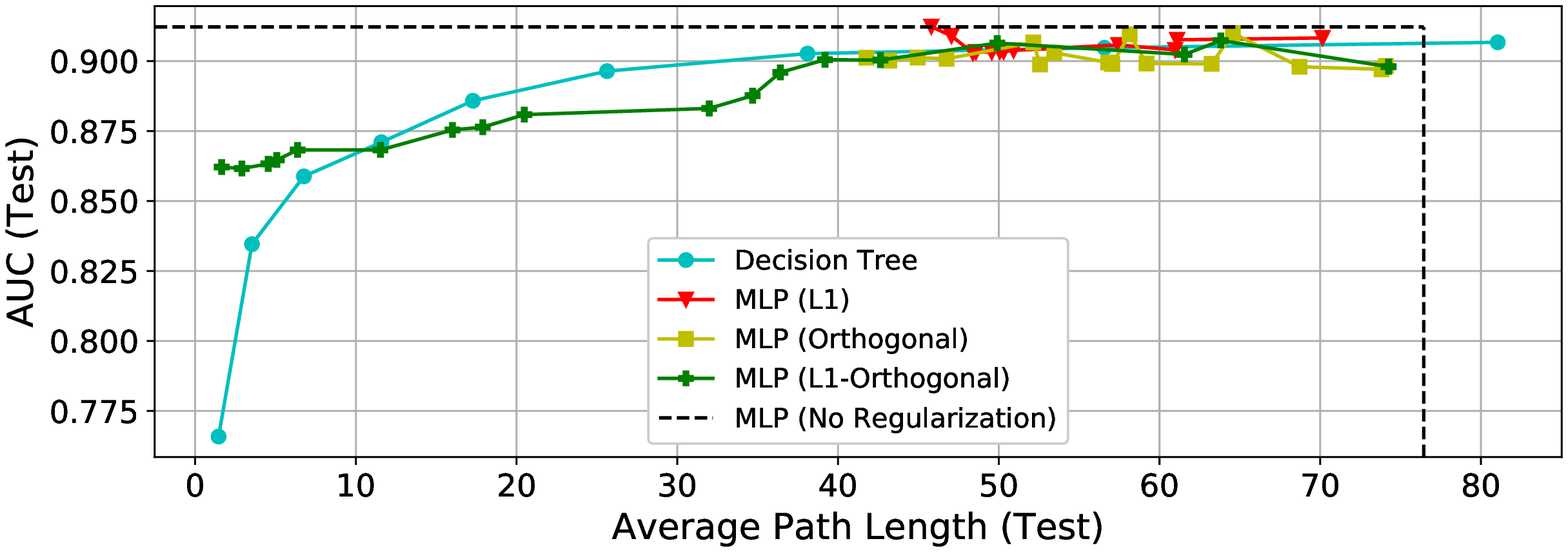}
		\vspace{-6mm}
		\caption{Adult}
	\end{subfigure}
	\begin{subfigure}[t]{0.49\textwidth}
		\centering
		\includegraphics[width=\textwidth]{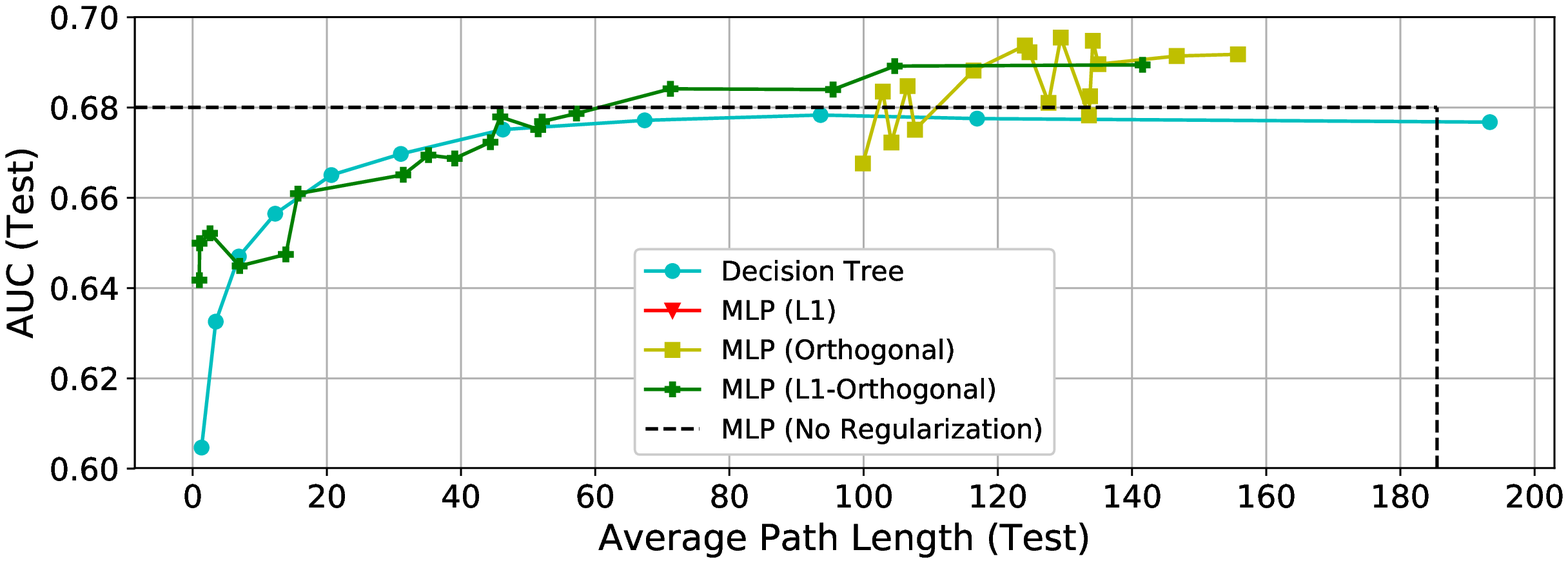}
		\vspace{-6mm}
		\caption{Diabetes}
	\end{subfigure}
	\caption[Fitness curves showing prediction quality vs. complexity.]{Fitness curves showing prediction quality (AUC) vs. complexity (APL).}
	\label{figure:apl-auc-plots}
	\vspace{-2mm}
\end{figure*}

In Fig.~\ref{figure:apl-auc-plots}, MLP complexity is plotted against predictive performance for a selection of MLPs---each trained with different regularization strengths being displayed as single points in a 2D fitness space. Additionally, the AUC of an unregularized MLP is illustrated as black dotted line in order to provide a baseline.

For the 2D-parabola data set, L$_1$ regularization does not produce DTs with both small node counts and good predictions. In contrast, the DTs resulting from orthogonal and \Lo-O regularization are of reduced complexity---with a relatively low predictive accuracy, though. The only approach that produces highly accurate results with an at the same time small APL is tree regularization. However, comparing tree regularization and \Lo-O regularization on the Iris data set already shows, that \Lo-O regularization can compete with tree regularization. Both regularizers encourage simple and accurate models.

When looking at the fitness curves of the remaining real-world data sets, it can be noticed that \Lo-O regularization consistently outperforms the other regularizers. The differences between L$_1$ regularization, standalone DTs and \Lo-O regularization are relatively small for some data sets (cf. Iris, Breast Cancer Wisconsin), while for other data sets, \Lo-O regularization achieves much higher reductions in complexity than standalone DTs and L$_1$ regularization (cf. Pima Indians Diabetes, Titanic, Mushroom).
With orthogonal regularization alone, for non of the real-world data sets a significant reduction in model complexity can be achieved. 

In summary, it can be said, that \Lo-O regularization's performance on the 2D-parabola data set is relatively poor, while on the other data sets the approach convinces with \emph{simple and accurate} models. This can be explained by the concept behind this approach. With \Lo-O regularization, MLPs with few connections that nevertheless cover a broad spectrum of features due to their orthogonal alignment can be learned. Since the 2D-parabola data set contains only two features, the variety of feature combinations is limited. The other data sets contain more features providing more combinatorial options that can be exploited. Additionally, one prerequisite for the \Lo-O approach to perform well is that the data set admits for decision boundaries that are more or less parallel to axis and orthogonal. This might be another explanation for the average performance on the 2D-parabola problem.

Furthermore, the results show that the \emph{combination} of L$_1$ and orthogonal regularization makes the \Lo-O regularizer powerful. Applying it leads to significantly better results than applying L$_1$ and orthogonal regularization individually.

\subsection{Fidelity}
For measuring fidelity, first the ``best'' performing DT from the exploration over different regularization parameters is determined. This corresponds to the DT with the smallest APL that reaches the predictive performance of an unregularized MLP. Next, each data set is split into training (60\%), test (20\%), and validation (20\%) using five different seeds together with scikit-learn's~\cite{Buitinck.2013} \texttt{train\_test\_split} function. With these data sets five runs are performed with the regularization parameters obtained from the best model being \emph{fixed}. Table~\ref{table:evaluation:fidelity} lists the mean fidelity 
and corresponding standard deviations for DTs extracted from MLPs with \Lo-O (\Lo-norm and FN), tree, and L$_1$ regularization as well as from an unregularized MLP for all data sets.

\begin{table*}[t]
	\center
	\caption{Fidelity values of DT predictions.}
	\label{table:evaluation:fidelity}
	\begin{tabular}{l|c|c|c|c|c}
		\multirow{2}{*}{\textbf{Data set}}
		& \multicolumn{2}{c|}{\textbf{\Lo-O}} & \multirow{2}{*}{\textbf{L1}} & \multirow{2}{*}{\textbf{Tree}} & \multirow{2}{*}{\textbf{unregularized}}\\ \cline{2-3} 
		& \textbf{\Lo-norm} & \textbf{FN}     &    &      &              \\ \hline
		2D-parabola             & 0.95\,$\pm$\,0.00 & 0.96\,$\pm$\,0.02  & \textbf{0.97}\,$\pm$\,0.02 & 0.96\,$\pm$\,0.01 & 0.96\,$\pm$\,0.01 \\
		Iris                    & \textbf{0.99}\,$\pm$\,0.02 & 0.97\,$\pm$\,0.02 & \textbf{0.99}\,$\pm$\,0.02 & 0.97\,$\pm$\,0.02 & 0.96\,$\pm$\,0.01 \\
		Breast Cancer Wisconsin & \textbf{0.95}\,$\pm$\,0.01 & 0.93\,$\pm$\,0.02 & 0.93\,$\pm$\,0.02 & \textbf{0.95} \,$\pm$\, 0.01 & 0.93\,$\pm$\,0.02\\
		Pima Indians Diabetes & \textbf{0.90}\,$\pm$\,0.01 & 0.88\,$\pm$\,0.02 & 0.87\,$\pm$\,0.03 & - & 0.86\,$\pm$\,0.03 \\
		Titanic                 & \textbf{0.94}\,$\pm$\,0.02 & \textbf{0.94}\,$\pm$\,0.02 & 0.93\,$\pm$\,0.01  & - & 0.89\,$\pm$\,0.02\\
		Mushroom                & \textbf{0.98}\,$\pm$\,0.00  & \textbf{0.98}\,$\pm$\,0.00 & \textbf{0.98}\,$\pm$\,0.00 & - & \textbf{0.98}\,$\pm$\,0.00 \\
		Adult                   & \textbf{0.95}\,$\pm$\,0.01  &\textbf{0.95}\,$\pm$\,0.00 & 0.95\,$\pm$\,0.01 & - & 0.94\,$\pm$\,0.00 \\
		Diabetes                   & 0.92\,$\pm$\,0.01  &\textbf{0.97}\,$\pm$\,0.02 & - & - & 0.81\,$\pm$\,0.01
	\end{tabular}
	\vspace{-2mm}
\end{table*}

Here, two things can be noticed. First, DTs extracted from regularized MLPs have very high fidelity scores that are mostly in a similar range. Second, regularization in general has a positive impact on fidelity, as with regularization the fidelity improves in most cases while simultaneously the APLs are reduced. This can be observed particularly for Diabetes and Pima Indians Diabetes data sets. In general, \Lo-O regularization achieves excellent fidelity scores: all but one score are $\geq0.90$. Thus, by analyzing the DTs it is possible to draw reliable conclusions about behavior and decision making of the MLP from which they were extracted.

\subsection{Comprehensibility}
\begin{figure}[tp]
	\centering
	\captionsetup{justification=centering}
	\begin{subfigure}[b]{0.3\textwidth}
		\centering
		\includegraphics[width=0.3\textwidth]{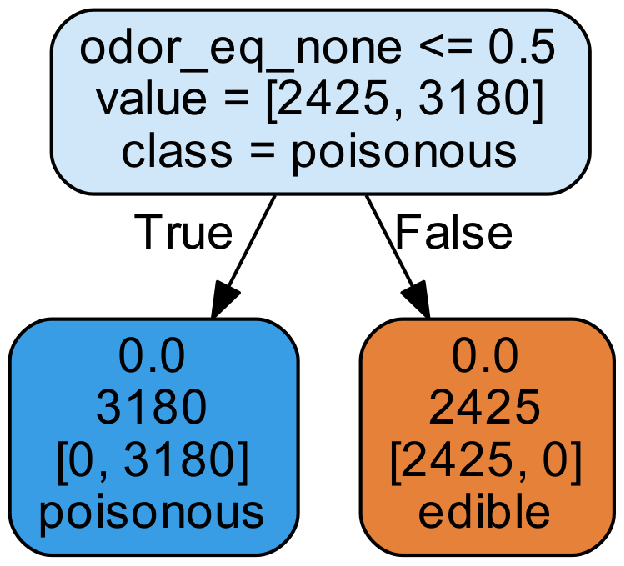}
		\caption{$\lambda_1=0.01, \lambda_{orth}=0.75$, AUC = 0.88}
		\label{fig:dts-mushroom_tiny}
		\vspace{2mm}
	\end{subfigure}
	\begin{subfigure}[b]{0.5\textwidth}
		\centering
		\includegraphics[width=.7\textwidth]{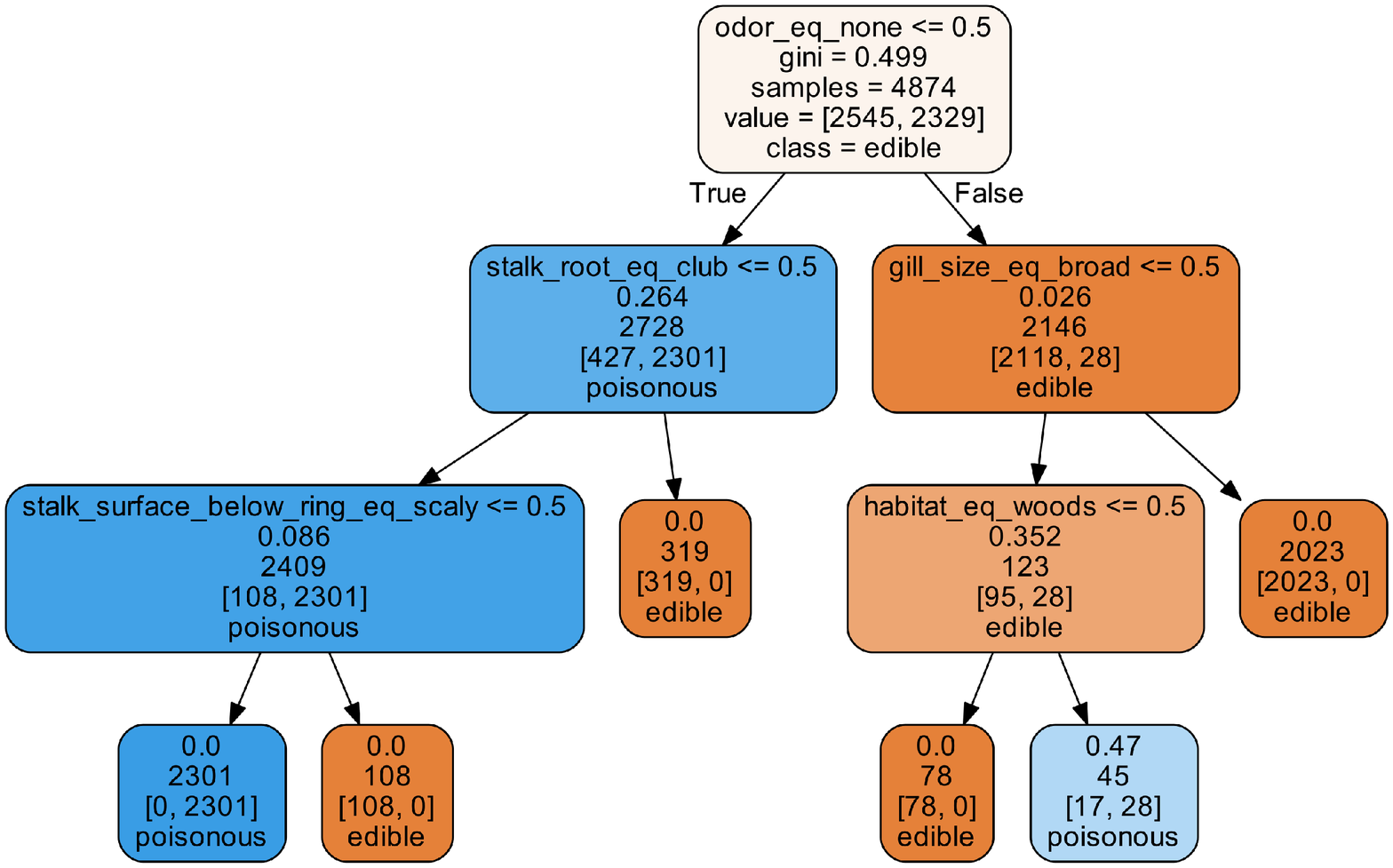}
		\caption{$\lambda_1=0.07, \lambda_{orth}=1.5$, AUC = 0.97 \newline}
		\label{fig:dts-mushroom_small}
		\vspace{-2mm}
	\end{subfigure}
	\caption{Small DTs for Mushroom data set.}
	\label{figure:dts-mushroom}
	\vspace{-2mm}
\end{figure}

Fig.~\ref{fig:dts-mushroom_tiny} and \ref{fig:dts-mushroom_small} show two small DTs for the Mushroom data set with an APL of 1 and 6.1, respectively. This comparison demonstrates that an extremely short path length is not necessarily beneficial for the comprehensibility of the DTs. The DT in Fig.~\ref{fig:dts-mushroom_tiny} classifies instances by only testing on one feature (\texttt{odor\_eq\_none}). Given this little information, it is hardly possible to draw meaningful conclusions about the global model. In contrast, the DT shown in Fig.~\ref{fig:dts-mushroom_small} has a higher APL, but nevertheless provides information in a clear and concise manner and enables a more detailed investigation. It shows, for example, which combination of features leads to a specific prediction and gives information about the major features---as features which appear near the tree's root node are more important than those near the leaf nodes.

This example shows that model complexity and comprehensibility are not necessarily the same. Consequently, when optimizing an MLP with regard to model complexity \emph{and} comprehensibility one has to find a trade-off between the two demands. Although a reduction in model complexity can improve comprehensibility significantly, a model that is too simple is scarcely meaningful.

\subsection{Consistency}
To measure consistency of \Lo-O regularization, ten MLPs are trained with fixed regularization parameters and random weight initialization. Then, a DT is extracted from each MLP and is used for classifying the test examples to calculate~\eqref{eq:consistency}. 

Consistency is quantified for three regularization configurations. The results for four data sets are listed in Table~\ref{table:evaluation:consistency} in ascending order of the complexity of the generated DTs (APL). There are various interesting observations: (1) The stronger \Lo-O regularization the higher is the fidelity, which is a highly desirable behavior. 
(2) The consistency tends to become lower with a stronger \Lo-O regularization. There are two explanations for this behavior. First, DTs are known to be quite unstable, because each split point depends on the parent split. Thus, if a different feature gets selected as test node, the whole tree changes~\cite{Molnar.2018}. Second, MLPs were initialized with random weights and thus, the trained MLPs are all different as \eqref{equ:loss} is a non-convex optimization problem. \Lo-O regularization seems to be more sensitive to initialization.

\begin{table*}
	\center
	\caption{Consistency of DT predictions.}
	\label{table:evaluation:consistency}
	\begin{tabular}{l|llc|cc||llc|cc}
    	\multirow{2}{*}{Data set} & \multicolumn{5}{c||}{\Lo-O} & \multicolumn{5}{c}{FN}\\ \cline{2-11}
    	& $\lambda_1$ & $\lambda_\mathrm{orth}$ & \textbf{APL} & \textbf{Consistency} & \textbf{Fidelity} & $\lambda_1$ & $\lambda_\mathrm{orth}$ & \textbf{APL} & \textbf{Consistency} & \textbf{Fidelity}\\ \hline
		Titanic & 0.006 & 0.025 & 2.4 & 0.70 & 0.98$\pm$\,0.03 & 0.02 & 0.75 & 1.7 & 1.00 & 0.99$\,\pm$\,0.00\\ 
		& 0.001 & 0.01 & 4.7 & 1.00 & 0.93$\,\pm$\,0.00 & 0.003 & 0.75 & 4.2 & 0.70 & 0.97$\,\pm$\,0.01\\
		& 0.001 & 0.001 & 7.9 & 1.00 & 0.91\,$\pm$\,0.01 & 0.002 & 0.001 & 8.0 & 0.94 & 0.93$\,\pm$\,0.01\\ \hline
		Mushroom  & 0.1 & 0.75 & 1.4 & 0.90 & 1.00$\pm$\,0.00 & 0.1 & 0.75 & 3.8 & 0.53 & 0.99$\pm$\,0.0\\ 
		& 0.08 & 1.1 & 6.4 & 0.93 & 0.99$\,\pm$\,0.00 & 0.08 & 1.1 & 6.4 & 0.52 & 1.00$\pm$\,0.00\\
		& 0.003 & 0.001 & 14 & 1.0 & 0.98$\,\pm$\,0.00 & 0.003 & 0.001 & 14 & 1.0 & 0.98$\,\pm$\,0.00\\ \hline
		Adult & 0.09 & 0.5 & 1.7 & 0.82 & 1.00\,$\pm$\,0.00 & 0.08 & 1.1 & 1.7 & 0.86 & 0.99\,$\pm$\,0.00\\
		& 0.02 & 0.25 & 16 & 0.88 & 0.98\,$\pm$\,0.00 & 0.04 & 0.75 & 13.6 & 0.97 & 0.99\,$\pm$\,0.00\\
		& 0.0025 & 0.1 & 39 & 0.92 & 0.95\,$\pm$\,0.00 &  0.003 & 0.25 & 37.2 & 0.86 & 0.97\,$\pm$\,0.01\\ \hline
		Diabetes & 0.04 & 0.75 & 3.9 & 0.84 & 1.00\,$\pm$\,0.00 & 0.03 & 1.0 & 3.4 & 0.79 & 0.99\,$\pm$\,0.00\\
		& 0.02 & 0.25 & 16 & 0.59 & 0.98\,$\pm$\,0.01 & 0.02 & 1.1 & 14.1 & 0.94 & 0.99\,$\pm$\,0.00\\
		& 0.0075 & 0.075 & 57 & 0.87 & 0.96\,$\pm$\,0.01 & 0.004 & 0.01 & 57.9 & 0.83 & 0.94\,$\pm$\,0.00\\
	\end{tabular}
	\vspace{-2mm}
\end{table*}

\subsection{Computational complexity}
As it was not possible to produce satisfying results for tree regularization for all data sets, only the run-times measured for 2D-parabola, Iris, and Breast Cancer data sets can be compared. 
As Table~\ref{tab:runtime} indicates, 
the run times of tree-regularized models are 30--70 times higher than those of an \Loo-regularized model.

\begin{table}[t]
    \centering
    \caption{Computation time in seconds.}
    \label{tab:runtime}
    \begin{tabular}{c|ccc}
        Regularization & 2D-parabola & Iris & Breast Cancer \\ \hline
        \Lo-O & \textbf{42.72} & \textbf{0.59} & \textbf{6.44} \\
        Tree & 2,841.56 & 17.78 & 471.49
    \end{tabular}
\end{table}

\section{Conclusions and Future Work}
In this paper, \Lo-O regularization was proposed for improving the extraction of decision trees from deep neural networks. 
This regularization approach influences the network's nonlinear decision boundary in such a way that it can be well-approximated by small DTs. These explainable models can provide users of different interest-groups insights into the decision making of the associated network in a representation format that is easily comprehensible. For example, one can derive important features as well as feature interdependencies by examining the DTs themselves. Furthermore, users can examine the DTs together with the input data in order to quickly retrace and \textit{simulate} what the complex model is doing.
It was shown that the extracted trees mimic the associated networks with high fidelity. Consequently, the conclusions drawn by interpreting the DTs are highly consistent in terms that they reflect the behavior of the associated deep model.

The experiments proved the technical ability of \Lo-O regularization to reduce model complexity. However, the inspection of the generated DTs showed that extremely short path lengths do not necessarily have to be favorable for DT comprehensibility. Thus, a next step could be to conduct detailed user studies with different user groups (including domain experts) in order to analyze the extracted DTs with regard to their comprehensibility and supportive potential.

This paper focused exclusively on the extraction of DTs from MLPs. 
Future work is also devoted to explore the effect of \Lo-O regularization on other types of explainable models like decision tables or decision sets. Also applicability of \Lo-O regularization for different NN types like recurrent or convolutional NNs needs to be investigated.

\section*{Acknowledgment}
This work was partially supported by the Ministry of Economic Affairs of the state Baden-W\"urttemberg (Zentrum f\"ur Cyber Cognitive Intelligence, Grant No. 017-192996) and by USU Software AG. We also would like to thank Philipp Wagner for supporting the execution of parts of the experiments.


\begin{thebibliography}{10}
\providecommand{\url}[1]{#1}
\csname url@samestyle\endcsname
\providecommand{\newblock}{\relax}
\providecommand{\bibinfo}[2]{#2}
\providecommand{\BIBentrySTDinterwordspacing}{\spaceskip=0pt\relax}
\providecommand{\BIBentryALTinterwordstretchfactor}{4}
\providecommand{\BIBentryALTinterwordspacing}{\spaceskip=\fontdimen2\font plus
\BIBentryALTinterwordstretchfactor\fontdimen3\font minus
  \fontdimen4\font\relax}
\providecommand{\BIBforeignlanguage}[2]{{%
\expandafter\ifx\csname l@#1\endcsname\relax
\typeout{** WARNING: IEEEtran.bst: No hyphenation pattern has been}%
\typeout{** loaded for the language `#1'. Using the pattern for}%
\typeout{** the default language instead.}%
\else
\language=\csname l@#1\endcsname
\fi
#2}}
\providecommand{\BIBdecl}{\relax}
\BIBdecl

\bibitem{Wang.2017}
T.~Wang, C.~Rudin, F.~Doshi-Velez, Y.~Liu, E.~Klampfl, and P.~MacNeille, ``{A
  Bayesian framework for learning rule sets for interpretable
  classification},'' \emph{{The Journal of Machine Learning Research}},
  vol.~18, no.~1, pp. 2357--2393, 2017.

\bibitem{Ribeiro.20160809}
M.~T. Ribeiro, S.~Singh, and C.~Guestrin, ``{``Why Should I Trust You?'':
  Explaining the Predictions of Any Classifier},'' in \emph{Proceedings of the
  22nd ACM SIGKDD International Conference on Knowledge Discovery and Data
  Mining}.\hskip 1em plus 0.5em minus 0.4em\relax New York, NY, USA: ACM, 2016,
  pp. 1135--1144.

\bibitem{Zilke.2016}
J.~R. Zilke, {Loza Menc{\'i}a E.}, and {Janssen Frederik}, ``{DeepRED -- Rule
  Extraction from Deep Neural Networks},'' in \emph{{Discovery Science}}, ser.
  {Lecture Notes in Computer Science}, T.~Calders, M.~Ceci, and D.~Malerba,
  Eds., 2016, pp. 457--473.

\bibitem{Molnar.2018}
C.~Molnar, \emph{{Interpretable Machine Learning. A Guide for Making Black Box
  Models Explainable}}, 2019,
  \url{https://christophm.github.io/interpretable-ml-book/}.

\bibitem{Bach.2015}
S.~Bach, A.~Binder, G.~Montavon, F.~Klauschen, K.-R. M{\"u}ller, and W.~Samek,
  ``{On Pixel-Wise Explanations for Non-Linear Classifier Decisions by
  Layer-Wise Relevance Propagation},'' \emph{{PloS one}}, vol.~10, no.~7, p.
  e0130140, 2015.

\bibitem{Friedman.2001}
J.~H. Friedman, ``{Greedy function approximation: A gradient boosting
  machine},'' \emph{{Ann. Statist.}}, vol.~29, no.~5, pp. 1189--1232, 2001.

\bibitem{Lakkaraju.2017}
H.~Lakkaraju, E.~Kamar, R.~Caruana, and J.~Leskovec, ``{Interpretable {\&}
  Explorable Approximations of Black Box Models},'' arXiv:1707.01154.

\bibitem{Puri.2017}
N.~Puri, P.~Gupta, P.~Agarwal, S.~Verma, and B.~Krishnamurthy, ``{MAGIX: Model
  Agnostic Globally Interpretable Explanations},'' arXiv:1706.07160.

\bibitem{Augasta.2011}
M.~G. Augasta and T.~Kathirvalavakumar, ``{Reverse Engineering the Neural
  Networks for Rule Extraction in Classification Problems},'' \emph{{Neural
  Processing Letters}}, vol.~28, no.~3, 2011.

\bibitem{Russell.2010}
S.~J. Russell and P.~Norvig, \emph{{Artificial intelligence. A modern
  approach}}, 3rd~ed., ser. {Prentice-Hall series in artificial
  intelligence}.\hskip 1em plus 0.5em minus 0.4em\relax Upper Saddle River, NJ,
  USA: {Pearson Education Inc.}, 2010.

\bibitem{Craven.1994.b}
M.~W. Craven and J.~W. Shavlik, ``{Extracing tree-structured representations of
  trained networks},'' in \emph{{Proceedings of the 8th International
  Conference on Neural Information Processing Systems (NIPS)}}, 1996, pp.
  24--30.

\bibitem{MikeWu.2018}
M.~Wu, M.~C. Hughes, S.~Parbhoo, M.~Zazzi, V.~Roth, and F.~Doshi-Velez,
  ``Beyond sparsity: Tree regularization of deep models for interpretability,''
  in \emph{AAAI}, 2018.

\bibitem{Xie.2017}
P.~Xie, B.~P{\'o}czos, and E.~P. Xing, ``{Near-Orthogonality Regularization in
  Kernel Methods},'' \emph{{Uncertainty in Artificial Intelligence}}, Aug.
  2017.

\bibitem{Xie.2017b}
P.~Xie, H.~Zhang, and E.~P. Xing, ``{Learning Less-Overlapping
  Representations},'' arXiv:1711.09300.

\bibitem{Boyd.2004}
S.~Boyd and L.~Vandenberghe, \emph{{Convex Optimization}}.\hskip 1em plus 0.5em
  minus 0.4em\relax New York, NY, USA: {Cambridge University Press}, 2009.

\bibitem{Andrews.1995}
R.~Andrews, J.~Diederich, and A.~B. Tickle, ``{Survey and critique of
  techniques for extracting rules from trained artificial neural networks},''
  \emph{{Knowledge-Based Systems}}, vol.~8, no.~6, pp. 373--389, 1995.

\bibitem{Kingma.2014}
D.~P. Kingma and J.~Ba, ``{Adam: A Method for Stochastic Optimization},''
  arXiv:1412.6980.

\bibitem{Fisher.1936}
R.~A. Fisher, ``{The Use of Multiple Measurements in Taxonomic Problems},'' in
  \emph{{Annals of Eugenics}}, 1936, vol.~7, pp. 179--188.

\bibitem{Dheeru.2017}
\BIBentryALTinterwordspacing
D.~Dua and C.~Graff, ``{{UCI} Machine Learning Repository},'' 2017. [Online].
  Available: \url{http://archive.ics.uci.edu/ml}
\BIBentrySTDinterwordspacing

\bibitem{KaggleInc..2012}
\BIBentryALTinterwordspacing
``{Titanic: Machine Learning from Disaster},'' 2012, last accessed 29 Sep 2019.
  [Online]. Available: \url{https://www.kaggle.com/c/titanic/data}
\BIBentrySTDinterwordspacing

\bibitem{Strack.2014}
B.~Strack, J.~P. DeShazo, C.~Gennings, J.~L. Olmo, S.~Ventura, K.~J. Cios, and
  J.~N. Clore, ``{Impact of HbA1c Measurement on Hospital Readmission Rates:
  Analysis of 70,000 Clinical Database Patient Records},'' \emph{{BioMed
  research international}}, 2014.

\bibitem{Buitinck.2013}
L.~Buitinck, G.~Louppe, M.~Blondel, F.~Pedregosa, A.~Mueller, O.~Grisel,
  V.~Niculae, P.~Prettenhofer, A.~Gramfort, J.~Grobler, R.~Layton,
  J.~Vanderplas, A.~Joly, B.~Holt, and G.~Varoquaux, ``{API design for machine
  learning software: experiences from the scikit-learn project},'' in
  \emph{{ECML PKDD Workshop: Languages for Data Mining and Machine Learning}},
  2013, pp. 108--122.

\end{thebibliography}

\end{document}